\documentclass{article}

\usepackage{microtype}
\usepackage{graphicx}
\usepackage{booktabs} 

\usepackage{hyperref}

\usepackage[accepted]{icml2023}

\usepackage{amsmath}
\usepackage{amssymb}
\usepackage{mathtools}
\usepackage{amsthm}

\usepackage[capitalize,noabbrev]{cleveref}

\usepackage{multirow}
\usepackage{subcaption}

\theoremstyle{plain}

\theoremstyle{definition}

\theoremstyle{remark}

\providecommand{\Paren}[1]{\ensuremath{\left( #1 \right)}}

\providecommand{\cL}{\mathcal{L}}
\def\tr{^\top}
\providecommand{\norm}[1]{\lVert#1\rVert}
\providecommand{\Norm}[1]{\ensuremath{\left\lVert#1\right\rVert}}
\newcommand{\ALGOCOMMENT}[2][.1\linewidth]{\leavevmode\hfill\makebox[#1][r]{//~#2}}
\newcounter{tablerownumbers}
\newcommand\rownumber{\stepcounter{tablerownumbers}\arabic{tablerownumbers}}

\usepackage[textsize=tiny]{todonotes}

\icmltitlerunning{Dual Propagation: Accelerating Contrastive Hebbian Learning with Dyadic Neurons}

\begin{document}

\twocolumn[
\icmltitle{Dual Propagation: \\
           Accelerating Contrastive Hebbian Learning with Dyadic Neurons}



\icmlsetsymbol{equal}{*}

\begin{icmlauthorlist}
\icmlauthor{Rasmus Høier}{cth}
\icmlauthor{D. Staudt}{cth}
\icmlauthor{Christopher Zach}{cth}
\end{icmlauthorlist}

\icmlaffiliation{cth}{Chalmers University of Technology, Sweden}

\icmlcorrespondingauthor{Rasmus Høier}{hier@chalmers.se}

\icmlkeywords{Machine Learning, ICML}

\vskip 0.3in
]



\printAffiliationsAndNotice{}  

\begin{abstract}
Activity difference based learning algorithms---such as contrastive Hebbian learning and equilibrium propagation---have been proposed as biologically plausible alternatives to error back-propagation. However, on traditional digital chips these algorithms suffer from having to solve a costly inference problem twice, making these approaches more than two orders of magnitude slower than back-propagation. In the analog realm equilibrium propagation may be promising for fast and energy efficient learning, but states still need to be inferred and stored twice.
Inspired by lifted neural networks and compartmental neuron models we propose a simple energy based compartmental neuron model, termed dual propagation, in which each neuron is a dyad with two intrinsic states. At inference time these intrinsic states encode the error/activity duality through their difference and their mean respectively. The advantage of this method is that only a single inference phase is needed and that inference can be solved in layerwise closed-form. Experimentally we show on common computer vision datasets, including Imagenet32x32, that dual propagation performs equivalently to back-propagation both in terms of accuracy and runtime.

\end{abstract}

\section{Introduction}
\begin{table*}[t]
\tabcolsep=0.11cm
\centering
\caption{Comparison of selected biologically motivated algorithms and back-propagation.}
\label{tab:method_comparison}
\begin{tabular}{|l|lll|lll|}
\hline
Global synchron. & \multicolumn{3}{c|}{Yes} & \multicolumn{3}{c|}{No} \\ \hline
Method & BP & CHL & EP & DCMC & LPOM & DP \\
States & Activity/error & Clamped/free & Nudged/Free & Activity/error & Clamped/Free & +/- nudged \\
Inference & Closed-form & Iterative & Iterative & Iterative & Iterative & Layerwise closed-form \\
Steps & 2 & (10--100) & $\sim300$ & $\sim1000$ & (10--100) & $\geq 2$  \\ \hline
\end{tabular}
\end{table*}

\begin{figure}[ht]
    \centering
     \begin{subfigure}[b]{0.23\textwidth}
         \centering
         \includegraphics[width=\textwidth]{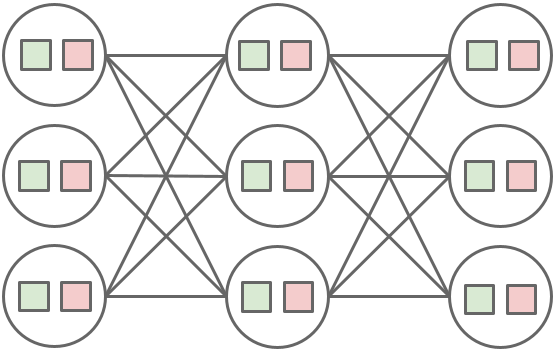}
         \caption{}
         \label{fig:dualprop_network}
     \end{subfigure}
     \begin{subfigure}[b]{0.23\textwidth}
         \centering
         \includegraphics[width=\textwidth]{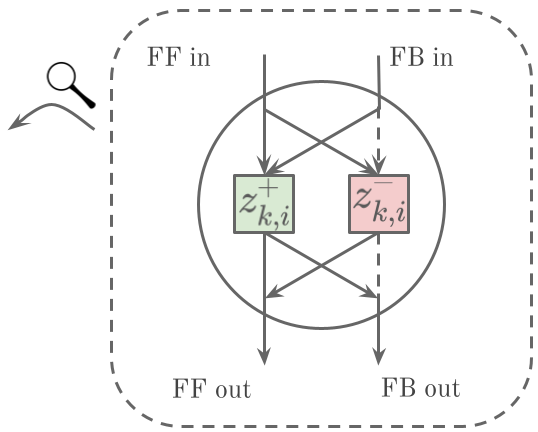}
         \caption{}
         \label{fig:dualprop_neuron}
     \end{subfigure}
     \caption{(a) A fully connected network with compartmental neurons. (b) Close-up of neuron $i$ in layer $k$. Solid arrows indicate identity connections and dashed arrows indicate sign inversion. The neuron possesses positively and negatively nudged states $z^+_{k,i}$ and $z^-_{k,i}$ and propagates two types of signals. The mean $\frac{1}{2}(z^+_{k,i} + z^-_{k,i})$ is propagated upstream to layer $k+1$ while the difference $\frac{1}{2}(z^+_{k,i} - z^-_{k,i})$ is propagated downstream to layer $k-1$.}
    \label{fig:diagram}
\end{figure}
In spite of the massive success of the error back-propagation (BP) method for training deep neural networks, there are several theoretical and practical reasons to consider alternatives.
One theoretical reason is the question of biological plausibility, which was already raised in~\cite{crick1989recent} soon after the publication of the influential back-propagation paper~\cite{rumelhart1986learning}.
A significant practical challenge is the energy consumption of using back-propagation to train deep neural networks~\cite{strubell2019energy}, which can be largely attributed to floating point processing but also to the required fine-grained synchronization of the individual steps---therefore preventing the utilization of e.g. energy efficient analog circuits~\cite{yi2022memristorcrossbars}.
Consequently, taking inspiration from biological motivated learning algorithms may contribute to making deep neural networks substantially more energy efficient.
Research along these lines has converged on the idea of encoding weight updates in terms of neuronal activity differences, Neural Gradient Representation by Activity Differences (NGRAD)~\cite{lillicrap2020NGRAD}. 
NGRAD methods differ in terms of what is meant by activity difference. Some approaches use the difference in neuron activity at different times, while others compute the difference between different sub-states within individual neurons.

Contrastive Hebbian learning~\cite{movellan1991,xie2003} and its modern incarnation~\cite{scellier2017} fall into the NGRAD category of methods and replace the fine-grained synchronization in BP with globally synchronizing two inference phases.
This allows continuous and asynchronous transmission of neural activity signals e.g.\ in (partially) analog computing devices (e.g.~\cite{oconnor2019training,zoppo2020equilibrium,kendall2020}).
Lifted neural networks (such as~\cite{carreira2014distributed,zhang2017convergent,li2020}) only require a single, largely asynchronously operating phase, but lack the simplicity in computing the neural activations that is found in back-propagation or in Hopfield nets.
This might be the factor preventing lifted networks---to our knowledge---being implemented in neuromorphic or analog devices.


In this paper we propose a single phase contrastive Hebbian learning-type algorithm, which we refer to as dual propagation (DP).
Neural units in our framework maintain internally two states (a ``dyad'') and are therefore an instance of a compartment based neuron model. 
The resulting weight update rules are fully local, and the inference rules have layerwise closed-form solutions.
Experimentally we explore the effects of different neuron state update schemes on MNIST, including one which updates layers of neurons in a random sequence as well as a resource efficient version which can be efficiently implemented on top of existing auto-differentiation frameworks. Furthermore, the resource efficient version is benchmarked on CIFAR10, CIFAR100 and Imagenet32x32, where it performs equivalently to back-propagation both in terms of accuracy and computational runtime.

Table~\ref{tab:method_comparison} lists a selection of NGRAD algorithms as well as back-propagation grouped into two groups based on whether they require global synchronization. The number of steps denotes how many state updates per neural unit are required. This information was not available for CHL and LPOM, hence we provide estimates based on our experience. For EP and DCMC the number of steps was based on numbers reported in~\cite{laborieux2021scaling, laborieux2022holomorphic} and \cite{sacramento2018} respectively. These numbers are dependent on network architecture, so they should be considered indicative only. 


\section{Related Work}\label{sec:related work}

\paragraph{Contrastive Hebbian learning and equilibrium propagation}\label{subsec:chl_and_eqprop}
Contrastive Hebbian learning was originally proposed for Hopfield networks with continuous units~\cite{movellan1991}, but has since been applied to layered networks as well~\cite{oreilly1996, xie2003}. In this framework inference consists in minimizing a Lyapunov function (or network potential) with respect to activations via suitable iterations.
During training this procedure is carried out twice (once with and once without clamping the output units to the target label), yielding so-called clamped and free steady states. Learning amounts to minimizing the difference between the Lyapunov function for these two states. 
The necessary global synchronization to initiate the two phases is somewhat problematic for a biologically motivated algorithm, as biological neural networks operate in a streaming context

More recently, equilibrium propagation~\cite{scellier2017, scellier2018} has been proposed as an improvement on CHL. In this framework output neurons are not clamped but instead ``nudged'' (or soft clamped via a target loss) towards the target labels.
Like CHL, equilibrium propagation is computationally costly when implemented on conventional hardware. However, this is not an issue when using equilibrium propagation for training analog neural networks~\cite{kendall2020, scellier2021thesis}.

\paragraph{Lifted networks}\label{subsec:lifted}
In the lifted neural network framework, the loss function is augmented by terms which penalize neurons not conforming to a chosen activation function~\cite{askari2018lifted}. Learning proceeds by first minimizing this augmented objective with respect to neuron activations and subsequently with respect to the weights. Different variations of this idea have been explored in~\cite{carreira2014distributed,zhang2017convergent,whittington2017approximation,gu2020fenchellifted, hoier2020, li2019, li2020, zach2019, song2020can}.
Lifted neural network and two-phase CHL are actually two sides of the same coin, as lifted networks implicitly integrate one of the two phases~\cite{zach2021}.

\paragraph{Compartmental neuron models}
The segregated dendrite model~\cite{Guerguiev2017} takes inspiration from pyramidal neurons, formulating inference and learning rules in terms of simplified dynamics between apical and basal dendritic compartments and the soma. However, similarly to CHL this model requires global synchronization of two distinct phases in order to compute errors as the temporal difference in apical dendritic activity.
An alternative to computing errors in terms of temporal activity differences is to encode errors as the activity difference between distinct neuronal states. This is the approach taken by dendritic cortical microcircuits (DCMC)~\cite{sacramento2018}. Hence the network does not require a global synchronization between distinct phases. However, this comes at the cost of requiring auxiliary neurons (interneurons). Note that both of these particular methods aim for a higher degree of biological plausibility, by modelling spiking neurons, which also makes them comparatively computationally costly.


\paragraph{Asynchronous DNN training}
A number of methods aim to improve the parallelism in training DNNs by decoupling the layer dependency.
Approaches proposed in the literature include auxililary coordindates (\cite{carreira2014distributed}, an early instance of a lifted network), decoupled training using ADMM~\cite{taylor2016training} and block-coordinate method~\cite{zhang2017convergent}, and synthetic gradients~\cite{jaderberg2017decoupled}.
Usually, these methods are not inspired by biological plausibility, but aim at better utilization of highly paralleized computing resources.

\paragraph{Weight transport}
While biological neural networks use distinct unidirectional pathways to transport signals to and from neurons artificial, neural networks trained with BP employs weight symmetry (activity is transported by $W$ and errors by $W\tr$).
Feedback alignment (FA)~\cite{lillicrap2014, lillicrap2016} and direct feedback alignment (DFA)~\cite{nokland2016}) are variations of back-propagation that aim to remove this symmetry constraint\footnote{Occasionally these algorithms are referred to as random back-propagation~\cite{baldi2018}}. In (D)FA errors are transported backwards using a distinct set of static randomly initialized weights. For fully connected architectures the learnable forwards weights can be observed to partially align to the static backwards weights, providing useful learning signals. However, deep convolutional networks trained with (D)FA fail to learn efficiently~\cite{bartunov2018, moskovitz2018}. Refenetti et al~\cite{refinetti2021} explains this in terms of the difficulty of having the highly sparse Toeplitz representation of a convolutional layers align to random feedback weights.
Difference target propagation~\cite{lee2015} also aim to address the weight transport problem, but take a different approach. In this framework each layer is modelled as an autoencoder, and distinct feed-back weights are trained to propagate useful targets to hidden layers by approximately inverting the feed-forward mapping. DTP has been explored in a variety flavors~\cite{lee2015, bartunov2018, meulemans2020, ernoult2022}, mainly differing in how feed-back weights are learned.

Although weight transport is not the focus of this paper, we explore a classic approach for training network with asymmetric feedforward and feedback weights. Kolen-Pollack learning~\cite{kolenpollack1994} of feedback weights amounts to applying the same weight updates to both sets of weights, combined with weight decay regularization. Given enough updates the weights will then converge to the same values. This approach has previously been used to facilitate learning in networks with distinct feedforward and feedback weights~\cite{Akrout2019, laborieux2021scaling}.
Weight mirrors~\cite{Akrout2019} and stochastic approximation to estimate the feedback mapping~\cite{ernoult2022} are recent contributions to address the weight transport problem.
We utilize the Kolen-Pollack scheme in some experiments to asses the general compatibility of our dual propagation method with enhanced biological plausibility.

    

\paragraph{Notations}
For a differentiable and strictly convex function $G$ we denote the regular Bregman divergence by $D_G(z\|y) = G(z)-G(y)-(z-y)\tr \nabla G(y)$ and its reparametrized version by $\bar D_G(z\|x) := G(z) - z\tr x + G^*(x)$. $\bar D_G(z\|x)$ is non-negative due to the Fenchel-Young inequality, and $\bar D_G(z\|x)=0$ iff $z=\nabla G^*(x) = f(x)$ for a suitable mapping $f=\nabla G^* = (\nabla G)^{-1}$. We therefore have $D_G(z\|g(x))=\bar D_G(z\|x)$. Constraints $x\in C$ are written as $\imath_C(x)$ in their functional form.

\section{Contrastive Hebbian Learning with Dyadic Neurons}

In this section we present a framework inspired by contrastive Hebbian learning, that is based on positively and negatively nudged internal states maintained for every neuron.
Consequently, the proposed framework shares a number of high-level similarities with other lifted neural network approaches, but retains closed-form inference steps to determine the neural activations.
Further, under suitable smoothness assumption on the activation non-linearities, the gold-standard BP gradient is approximated to second order.


\subsection{The Contrastive Objective}

Our model is based on two sets of states for neural activity denoted by $z^+$ and $z^-$. Since we focus on layered feed-forward DNNs, $z_k^\pm$ denotes the respective activations in layer $k$, where layer $0$ is the input and layer $L$ is the output layer.
For a target loss $\ell$ the main objective in our model driving the update of the network parameters is given by
\begin{align}
\begin{split}
    &\cL_\alpha(\theta) := \min_{z^+} \max_{z^-} \alpha\ell(z_L^+) + \bar\alpha\ell(z_L^-) \nonumber \\
    &+ \sum_{k=1}^L \tfrac{1}{\beta_k}\! \Paren{ \bar D_{G_k}(z_k^+ \| W_{k-1}\bar z_{k-1}) \!-\! \bar D_{G_k}(z_k^- \| W_{k-1}\bar z_{k-1})}
    \nonumber 
\end{split}\\
\begin{split}
    &= \min_{z^+} \max_{z^-} \alpha\ell(z_L^+) + \bar\alpha\ell(z_L^-) \\
    &+ \sum_{k=1}^L \tfrac{1}{\beta_k}\! \Paren{ G_k(z_k^+) - G_k(z_k^-) + (z_k^- \!-\! z_k^+)\tr W_{k-1}\bar z_{k-1} },
    \label{eq:L_alpha}
\end{split}
\end{align}
where $\alpha\in[0,1]$, $\bar\alpha=1-\alpha$, $\bar z_k:= \alpha z_k^+ + \bar\alpha z_k^-$. $G_k$ are strictly convex and differentiable functions determining the activation non-linearity at layer $k$.
There are implicit constraints fixing $z_0^+=z_0^-=x$ to the network's input $x$, and the target loss $\ell$ carries the information on the desired prediction (e.g.\ target label).
For notational brevity we omit an explicit indication of bias vectors.

The general motivation for this cost function is that the states $z^+$ are ``nudged'' towards reducing the target loss $\ell$, and $z^-$ are negatively nudged states increasing the target loss.
Both states are regularized via the Bregman divergences in $\cL_\alpha$. An important property of~\eqref{eq:L_alpha} is that the problematic term $G_k^*(W_{k-1}z_{k-1})$ in $\bar D_{G_k}(z_k,W_{k-1}z_{k-1}$ cancels---at the expense of a min-max inner problem structure.
After the activations $z^\pm$ are determined in the inference phase, the network weights $\theta=\left(W_0,\dotsc,W_{L-1}\right)$ can be adjusted (see Section~\ref{sec:inference}).
The objective in~\eqref{eq:L_alpha} is stated for a single training sample $(x,\ell)$ but is straightforwardly extended over an entire training set.

The choice of $G_k$ determines the induced network non-linearity, e.g.\ $G_k=\norm{\cdot}^2/2$ yields linear units, $G_k=\norm{\cdot}^2/2+\imath_{\ge 0}(\cdot)$ introduces ReLU activation mappings (e.g.~\cite{zhang2017convergent}), and $G_k$ being the negated Shannon entropy adds a soft-max layer (e.g.~\cite{zach2021}).

It turns out that only three choices for $\alpha$ are particularly meaningful. Before focusing on the main case $\alpha=1/2$ in the remainder of this work, we briefly discuss the choices of $\alpha=1$ and $\alpha=0$.

\paragraph{The case $\alpha=1$:}
In this case it is possible to maximize w.r.t.\ $z^-$ in closed form, and the objective is given by
\begin{align}
    \cL_1(\theta) = \min_{z^+} \ell(z_L^+) + \sum_{k=1}^L \tfrac{1}{\beta_k} \bar D_{G_k} (z_k^+ \| W_{k-1} z_{k-1}^+),
\end{align}
which can be identified essentially as the LPOM objective proposed in~\cite{li2019}, which in the limit $\beta_k\to 0^+$ yields a back-propagation variant based on appropriate directional derivatives~\cite{zach2021}.
The optimal state $z^+$ minimizes the target loss, but is regularized to stay close to the forward pass prediction $f_k(W_{k-1} z_{k-1}^+)$.
One advantage of $\cL_1$ over similar lifted network formulations such as MAC~\cite{carreira2014distributed} is, that the inner minimization problem w.r.t.\ $z^+$ in $\cL_1$ is at least layer-wise convex, and inferring the network activations $z_k^+$ can be conducted by approximately solving
\begin{align}
\begin{split}
   \min_{z_k^+} &\,\beta_k \Paren{ G_k(z_k^+) - (z_k^+)\tr W_{k-1}z_{k-1}^+ } \\
   &+ \beta_{k+1} \Paren{ G_{k+1}^*(W_k z_k^+) - (z_{k+1}^+)\tr W_k z_k^+}.
\end{split}
\label{eq:LPOM_inference}
\end{align}
In general, determining $z_k^+$ (with all other states fixed) requires an iterative algorithm and cannot be obtained in closed-form.

\paragraph{The case $\alpha=0$:}
Similarly, by first using the min-max lemma and by minimizing w.r.t.\ $z^+$ in closed form, the resulting objective is an upper bound of
\begin{align}
    \cL_0(\theta) \ge \max_{z^-} \ell(z_L^-) - \sum_{k=1}^L \tfrac{1}{\beta_k} \bar D_{G_k} (z_k^- \| W_{k-1} z_{k-1}^-).
\end{align}
The r.h.s.\ can be interpreted as an ``anti-LPOM'' objective, and the solution for $z^-$ maximizes $\ell$, but is regularized by the second part in the loss.
Assuming $\min g(u)=0$ and $v:= \arg\min_u g(u)$, the general relation
\begin{align}
    \min_u f(u) \!+\! g(u) & \le f(v) \le \max_u f(u) \!-\! g(u)
\end{align}
implies that $\cL_0(\theta) \ge \cL_1(\theta)$.
Loosely speaking, $\cL_0$ uses the opposite directional derivatives (in the direction of the increasing loss in contrast to $\cL_1$) when $\beta_k\to 0^+$.
$\cL_0$ shares the block-concave structure of the inner tasks with layer-wise convexity of $\cL_1$, but also its difficulty of inference similar to~\eqref{eq:LPOM_inference}.
$\cL_0$ also puts an upper limit on the choice of $\beta_L$ in order to prevent an unbounded maximization instance w.r.t.\ $z_L^-$.

\paragraph{The general case $\alpha\in(0,1)$}
For any choice $\alpha\in(0,1)$ optimization over one set of unknowns (i.e.\ maximizing out $z^-$ or minimizing out $z^+$) is not easily possible in $\cL_\alpha$. Setting $\alpha\in(0,1)$ is nevertheless formally appealing, as layer-wise inference (Section~\ref{sec:inference}) is very efficient and can be conducted in closed-form.

Unlike contrastive Hebbian learning frameworks adapted for layered networks (such as~\cite{xie2003, zach2019}), the objective $\cL_\alpha$ in~\eqref{eq:L_alpha} does not need layer-wise discounting by using an increasing sequence for $\beta_k$ satisfying $\beta_k \ll \beta_{k+1}$. Generally, the most important parameter is $\beta_L$ determining the amount of ``nudging'' (or soft-clamping) introduced by the target loss $\ell$, and setting $\beta_k=\beta_L$ for all $k=1,\dotsc,L-1$ is sufficient in practice.

\subsection{Inference Rules and Weight Updates}
\label{sec:inference}

In this section we discuss the inference method to determine the solutions $z^+$ and $z^-$ of the inner optimization tasks in~\eqref{eq:L_alpha}.
The method is formally based on block-coordinate descent (BCD) by solving for $z_k^+$ and $z_k^-$ for a specific layer while keeping all other states $z_{\setminus k}^\pm$ fixed.
We emphasize that these steps are inspired by BCD, but due to the min-max structure convergence results for BCD are not readily applicable, and therefore the proposed inference method requires a different analysis (see Section~\ref{sec:analysis}).

Minimization and maximization over $z_k^+$ and $z_k^-$, respectively, in~\eqref{eq:L_alpha} can be carried out simultaneously for an entire layer, yielding the closed form inference rules
\begin{align}
\label{eq:zk_pm}
\begin{split}
    z_k^+ &\gets f_k \Paren{ W_{k-1} \bar z_{k-1} + \tfrac{\alpha\beta_k}{\beta_{k+1}} W_k\tr (z_{k+1}^+ - z_{k+1}^-)} \\
    z_k^- &\gets f_k \Paren{ W_{k-1} \bar z_{k-1} - \tfrac{\bar\alpha\beta_k}{\beta_{k+1}} W_k\tr (z_{k+1}^+ - z_{k+1}^-)}
\end{split}
\end{align}
for $k=1,\dotsc,L-1$. Recall that $z_0^+=z_0^-=x$ for the network input $x$, and that $\bar z_k := \alpha z_k^+ + \bar\alpha z_k^-$ is the possibly weighted average.
The assignment of $z_L^\pm$ depends on the target loss $\ell$. If the output layer is linear and $\ell(z_L) = \norm{z_L-y}^2/2$ is the least-squares loss, then we obtain
\begin{align}
\label{eq:zL_MSE}
\begin{split}
    z_L^+ &\gets \tfrac{1}{1 + \alpha \beta_L} \Paren { W_{L-1} \bar z_{L-1} + \alpha \beta_L y } \\
    z_L^- &\gets \tfrac{1}{1 - \bar\alpha \beta_L} \Paren{ W_{L-1} \bar z_{L-1} - \bar\alpha \beta_L y }.
\end{split}
\end{align}
Note that solving for $z_L^-$ is an unbounded convex maximization problem if $\bar\alpha \beta_L \ge 1$, and therefore we need to choose $\beta_L < 1/\bar\alpha = 1/(1\!-\!\alpha)$.
If the target loss is linear (or as occurring more commonly, a differentiable target loss is linearized), i.e.\ $\ell(z)=g\tr z$, then the update for a linear output layer reduces to
\begin{align}
\label{eq:zL_linear}
    z_L^+ \!\gets\! W_{L-1} \bar z_{L-1} \!-\! \alpha \beta_L g & & z_L^- \!\gets\! W_{L-1} \bar z_{L-1} \!+\! \bar\alpha \beta_L g.
\end{align}
%
It is worth noting that in the absence of upstream activity (or a constant target loss) the update in~\eqref{eq:zk_pm} basically reduces to a standard feed-forward pass, with $z_k^+=z_k^-$.

Once the states $z^\pm$ are inferred (or sufficiently close to their optimal solution), the contribution of a sample $(x,\ell)$ to the overall gradient w.r.t.\ the trainable parameters $\theta=(W_0,\dotsc,W_{L-1})$ is given by
\begin{align}\label{eq:dW}
    \tfrac{\partial}{\partial W_{k-1}} \cL_\alpha = \tfrac{1}{\beta_k}\left(z^-_{k} - z^+_{k}\right) \bar z_{k-1}\tr.
\end{align}
This is an instance of contrastive Hebbian learning using only information that is local to the artificial synapse.


\subsection{Analysis}
\label{sec:analysis}

In our analysis we consider only the case $\alpha=1/2$, i.e.\ $\cL_{1/2}$, since $\cL_1$ is discussed in the literature~\cite{li2019,li2020,zach2021} (and $\cL_0$ is quite related as discussed above), and the validity of the closed-form updates in~\eqref{eq:zk_pm} hinges on the choice $\alpha=1/2$ as described below in the convergence analysis.

\paragraph{Equivalence to back-propagation in the limit}
In this paragraph we demonstrate that the first factor in~\eqref{eq:dW}, $(z_k^--z_k^+)/\beta_k$, converges to
\begin{align}
    \tfrac{d}{dz_k} \ell(z_L) \qquad \text{s.t. } z_k = f_k(W_{k-1}z_{k-1})
\end{align}
i.e.
\begin{align}
\begin{split}
   \tfrac{d}{dz_L} \ell(z_L) &= \ell'(z_L) \\
   \tfrac{d}{dz_k} \ell(z_L) &= W_k\tr f_{k+1}'(W_k z_k)\, \tfrac{d}{dz_{k+1}} \ell(z_L)
\end{split}
\end{align}
for $\beta_k\to 0^+$. Consequently, the learning rule in~\eqref{eq:dW} approaches the back-propagation method in the limit.

We use $a_{k}:=W_{k-1}\bar z_{k-1}$ and $\Delta_k := \tfrac{1}{2\beta_k} (z_{k}^- - z_{k}^+)$ in the following.
Since we are interested in the limit case when $\beta_k\to 0^+$, it is sufficient to employ a linearized target loss, $\ell(z_L)=g\tr z_L$.
Using~\eqref{eq:zL_linear} we deduce that $(z_L^--z_L^+)/\beta_L = g$ and the claim is therefore true for $d\ell(z_L)/dz_L$.
For $k<L$ we expand
\begin{align}
\begin{split}
    & \lim_{\beta_k\to 0^+} \tfrac{z_k^--z_k^+}{\beta_k} \\
    &= \lim_{\beta_k\to 0^+} \tfrac{f(a_k + \beta_k W_k\tr \Delta_{k+1}) - f(a_k - \beta_k W_k\tr \Delta_{k+1})}{\beta_k} \\
    &= 2 W_k\tr f_{k+1}'(a_k) \Delta_k =  W_k\tr f_{k+1}'(a_k)\, \tfrac{z_{k+1}^- - z_{k+1}^+}{\beta_{k+1}},
\end{split}
\end{align}
but the last factor equals $d\ell(z_L)/dz_{k+1}$ by our induction hypothesis.
Hence, in the weak feedback setting (where $\beta_k \approx 0$ for all $k \in \{1,\dotsc,L\}$), dual propagation approximates back-propagation.
This property is not surprising and is also shared with many (contrastive) Hebbian learning frameworks.

This part of the analysis is valid for $\alpha\in[0,1]$ in general, as deviating from $\alpha=1/2$ only introduces an asymmetry in the finite differences estimate for the derivative.
The induction to show this equivalence requires $z_k^\pm$ to be fixed points of the updates in~\eqref{eq:zk_pm} and~\eqref{eq:zL_linear} to be applicable.
Consequently, we focus our attention on the question whether the state updates reach such fixed point at all in the remainder of this section.


\paragraph{Convergence analysis: linear networks}
We assume $G_k(z_k)=\norm{z_k}^2/2$ and a linearized loss, $\ell(z_L)=g\tr z_L$. For brevity we assume all $\beta_k$ are equal to a common $\beta>0$. In this setting the updates for $z_k^+$ and $z_k^-$ reduce to
\begin{align}
\begin{split}
    z_L^\pm &\gets W_{L-1} \bar z_{L-1} \mp \beta g \\
    z_k^\pm &\gets W_{k-1} \bar z_{k-1} \pm \tfrac{1}{2} W_k\tr (z_{k+1}^+ - z_{k+1}^-) \\
\end{split}
\end{align}
We reparametrize the updates in terms of
\begin{align}
  \bar z_k = \tfrac{1}{2} (z_k^+ + z_k^-) & & \delta_k = \tfrac{1}{2} (z_k^+ - z_k^-),
\end{align}
i.e., $z_k^+ = \bar z_k + \delta_k$ and $z_k^- = \bar z_k - \delta_k$. After rearranging terms, the update steps above translate to
\begin{align}
  \bar z_k \gets W_{k-1} \bar z_{k-1} & & \delta_L \gets -\beta g & & \delta_k \gets W_k\tr \delta_{k+1}
\end{align}
in terms of the average state $\bar z_k$ and error signal $\delta_k$. These steps can be identified as steps in the forward and backward pass of back-propagation in a linear network. A difference to regular back-propagation is that the layers are not traversed in a predefined order, but potentially in an arbitrary sequence.
The relevant observation is that $\bar z_k$ has the true value (and remains at that state) if the sequence of layer update contains the ordered sequence $[1:k]$ as subsequence. Analogously, $\delta_k$ is assigned (and fixed) to the correct error signal if $[L:k]$ is a subsequence of the traversed layers. Thus, the states $\bar z$ and $\delta$ (and consequently $z^+$ and $z^-$) have their correct values once the sequence of visited states contains one entire forward and backward pass as subsequences.
Hence, the condition on the traversal sequence is that $(1,\dotsc,L,\dotsc,1)$ appears eventually as a subsequence.
This is actually a necessary condition for all types of supervised learning methods for DNNs: the input needs to reach the loss, and the loss needs to be distributed through the entire network.

\paragraph{Convergence analysis: nonlinear networks}
We use similar ideas as above to analyze the proposed dual propagation method for nonlinear networks.
The activations resulting from the pure forward pass are denoted as $z^*$, i.e.\ $z_k^*=f_k(W_{k-1}z_{k-1}^*)$.
As before, we assume $\beta_k=\beta$ for all $k$ as above for notational brevity and assume a linearized target loss $\ell(z_L)=g\tr z_L$.
We use two back-propagated error signals $\delta^+$ and $\delta^-$ corresponding to forward and backward finite differences, respectively.
The underlying recursion for $\delta^\pm_k$ is $\delta^\pm_L \gets -\beta g$ and
\begin{align}
\begin{split}
  \delta_k^+ &\gets f_k\Paren{ W_{k-1}z_{k-1}^* + W_k\tr \delta_{k+1}^+ } - z_k^* \\
  \delta_k^- &\gets z_k^* - f_k\Paren{ W_{k-1}z_{k-1}^* - W_k\tr \delta_{k+1}^+ }
\end{split}
\label{eq:delta_k}
\end{align}
When $\beta \to 0^+$ and differentiable $f_k$ this approaches
\begin{align}
    \tfrac{1}{\beta} \delta_k^\pm \to f_k'(W_{k-1}z_{k-1}^*) W_k\tr \delta_{k+1}
\end{align}
(recall that $\delta_{k+1}^\pm$ scales with $\beta$ via $\delta_L^\pm=-\beta g$).
If $f_k$ is not differentiable at its argument, we obtain directional derivatives instead (if they exist).
As argued above, any sequence of updates $z_k^*\gets f_k(W_{k-1}z_{k-1}^*)$ and $\delta_k^\pm$ in~\eqref{eq:delta_k} eventually yields the correct forward activations and error signals (i.e.\ the finite-difference approximation of the derivative) under mild conditions.
After introducing $z_k^\pm =z_k^* \pm \delta_k^\pm$ we deduce the respective updates
\begin{align}
\begin{split}
  z_k^+ &= z_k^* + \delta_k^+ \gets f(W_{k-1}z_{k-1}^* + W_k\tr \delta_{k+1}^+) \\
  z_k^- &= z_k^* - \delta_k^- \gets f(W_{k-1}z_{k-1}^* - W_k\tr \delta_{k+1}^-),
\end{split}
\label{eq:zk_pm_update_nonlinear}
\end{align}
which are almost the updates given in~\eqref{eq:zk_pm}.
In general we need to maintain 3 quantities (e.g.\ $z_k^+$, $z_k^-$ and $z_k^*$, unless $\delta_k^+=\delta_k^-$).
By observing that
\begin{align}
  \bar z_k &= \tfrac{1}{2} (z_k^++z_k^-) = z_k^* + \tfrac{1}{2} (\delta_k^+ - \delta_k^-)
\end{align}
we conclude that $\bar z_k$ is a good approximation of $z_k^*$ as long as $\delta_k^+ \approx \delta_k^-$, which is satisfied if the activation function $f_k$ is at least locally approximately linear.
Consequently the algorithm replaces $z_{k-1}^*$ with $\bar z_{k-1}$ in~\eqref{eq:zk_pm_update_nonlinear} to maintain only two sets of neural activations (and to be in line with other CHL-based methods), thereby yielding the updates in~\eqref{eq:zk_pm}.
As shown in Section~\ref{sec:asymmetric}, this line of reasoning is applicable only when $\alpha=1/2$.
In toy experiments, choosing $\alpha$ significantly different from $1/2$ leads to at least inferior results (when all $\beta_k$ are small) or even to divergent behavior of $z^\pm$ when using~\eqref{eq:zk_pm} for, e.g., with the choice of $\beta_k=1/2$ all $k$. The case $\alpha=1/2$ works equally well for a wide range of $\beta_k<1/(1-\alpha)=2$.


It is interesting to note that learning the network parameters and the initial motivations for the state updates in~\eqref{eq:zk_pm} are based on~\eqref{eq:L_alpha}, but the state updates in~\eqref{eq:zk_pm_update_nonlinear} (and their approximation in~\eqref{eq:zk_pm}) are best understood as proceeding towards the global optimum of the following objective,
\begin{align}
\begin{split}
    &U(z^*,\delta^\pm) := \beta g\tr (\delta_L^+ + \delta_L^-) + \tfrac{1}{2} \norm{\delta_L^+}^2 + \tfrac{1}{2} \norm{\delta_L^-}^2 \\
    &+ \sum \Norm{\delta_k^+ - f_k\!\Paren{ a_k^* + W_k\tr \delta_{k+1}^+ } }^2 \\
    &+ \sum \Norm{\delta_k^- - f_k\!\Paren{ a_k^* - W_k\tr \delta_{k+1}^+ } }^2 + \sum \!D_{\bar G_k}(z_k^* \| a_k^*)
\end{split}
\label{eq:U}
\end{align}
(where $a_k^* := W_{k-1}z_{k-1}^*$ is the pre-synaptic activation). The last two lines are zero for the solution of the forward and backward pass, and $\delta_L^\pm=-\beta g$ minimizes the remaining first line.
The updates are not necessarily decreasing~\eqref{eq:U} in each step, in particular they cannot be derived as a block-coordinate method to minimize~\eqref{eq:U}.

\subsection{Biological plausibility}
One may conceptualize our framework in different ways. Mathematically it is natural to frame the learning objective in terms of the positively and negatively nudged states $z^+$ and $z^-$ as they are elements in the same space, and the optimization problem is easy to (approximately) solve. However, in a biological circuit one might also conceive of the difference ($\delta_k$) and mean ($\bar{z}_k$) of the nudged states as the actual compartments of the neuron. 

This is superficially similar to the approach taken by \cite{Guerguiev2017}, in which feedback signals are integrated in the apical dendrite and feedforward signals are integrated in the basal dendrite. However, in our case the error $\delta_k=\frac{1}{2}(z_k^+-z_k^-)$ also depends on the feed-forward input. Inspired by \cite{sacramento2018} this dependence on both feed-forward and feed-back signals (or a relaxed version of it), may be achieved using auxiliary neurons.

In practice our method (as well as all the other methods listed in Table~\ref{tab:method_comparison}) are far from modelling the complexity of biological neurons. Neuroscience research suggests that dendrites are multi-compartmental structures, which in terms of computational capabilities are closer to small networks of artificial neurons than to individual artificial neurons. E.g. a single dendrite can solve the XOR problem~\cite{chavlis2021bio_dendrites_2_ANNs, beniaguev2021single_cortical_column}, whereas a dot product based artificial neuron (i.e.~McCulloch-Pitts neuron) can not. As such the proposed dyadic framework is still a greatly simplified model, and should be seen as an example of a minimal circuitry required to perform effective credit-assignment while obeying certain known biological constraints, such as using only local information, being capable of operating asynchronously and not requiring computing the derivatives of the nonlinearities.

\section{Implementation}
We conducted all experiments by jointly optimizing $\cL_{1/2}$ in~\eqref{eq:L_alpha} with respect to weights and activations.
%
In order to train a network in a supervised setting using~\eqref{eq:zk_pm} and~\eqref{eq:dW}, information about the input data as well as about the prediction error needs to be propagated to all layers. In a network with $L$ layers (counting hidden layers and output layer) this would require $L-1$ updates of all neurons. However, since~\eqref{eq:zk_pm} has closed form solution, the most economical approach is to simply update neurons 
sequentially: layer $1\rightarrow2\rightarrow\dotsc\rightarrow L\rightarrow L-1\rightarrow\dotsc\rightarrow1$. 
If the network is initialized with zero activity, then the first $L-1$ updates reduce to a standard feed-forward pass. This is shown on the example of a single batch in Algorithm~\ref{alg:dualprop} (black and {\color[HTML]{0077BB}blue} lines). The algorithm can be efficiently implemented on top of an existing autodiff framework using custom derivative rules. 
A less cost-efficient approach (on traditional computing hardware) is to randomly select layers to update for some  $T_{max}$ number of iterations. This is done in a very similar way, as shown in Algorithm~\ref{alg:dualprop} by the {\color[HTML]{CC3311} red} lines (replacing the {\color[HTML]{0077BB}blue} ones).


\begin{algorithm}
  \caption{Dual propagation}\label{alg:dualprop}
  \begin{algorithmic}[1]
    \STATE $\mathbf{z^+_0}, ~\mathbf{z^-_0} \gets \mathbf{x}, ~\mathbf{x}$ \label{op0}
    \STATE $\mathbf{z^+_k}, ~\mathbf{z^-_k} \gets \mathbf{0}, ~\mathbf{0}\qquad \forall~0 < k \leq L$ \label{op1}
    {\color[HTML]{0077BB}\STATE\algorithmicfor~{$k\in [1,...,L-1]$}~\algorithmicdo\ALGOCOMMENT{Regular}} 
        {\color[HTML]{0077BB}\STATE\hspace{1em}$z_k^+$, $z_k^- \gets f_k(z_{k-1}^+)$, $f_k(z_{k-1}^-)$}
    {\color[HTML]{0077BB}\STATE \algorithmicend}
    {\color[HTML]{CC3311}\STATE\algorithmicfor~{$T\in [1,...,T_{max}]$}~\algorithmicdo\ALGOCOMMENT{Random}} 
        {\color[HTML]{CC3311}\STATE\hspace{1em}$k\gets \textrm{sample~inter~from~[1,..., L]}$} 
        {\color[HTML]{CC3311}\STATE\hspace{1em} update $z_k^+$, $z_k^-$  using~\eqref{eq:zk_pm} /~\eqref{eq:zL_MSE} /~\eqref{eq:zL_linear} }
    {\color[HTML]{CC3311}\STATE \algorithmicend}
    \FOR{$k\in [L,...,1]$}
        {\color[HTML]{0077BB}\STATE update $z_k^+$, $z_k^-$  using~\eqref{eq:zk_pm} /~\eqref{eq:zL_MSE} /~\eqref{eq:zL_linear} \ALGOCOMMENT{Regular}}
        \STATE $W_{k-1} \gets W_{k-1} - \frac{\eta}{2\beta_{k}}(z^+_{k} \!-\! z^-_{k})(z^+_{k-1} \!+\! z^-_{k-1})\tr$
    \ENDFOR
  \end{algorithmic}
\end{algorithm}

Our networks use the ReLU non-linearity, hence $G_k(z_k)=\norm{z_k}^2/2 + \imath_{\ge 0}(z_k)$ for $k=1,\dotsc,L-1$, and the output layer is linear ($G_L(z_L)=\norm{z_L}^2/2$).
Note that $G_k$ is not actually used in~\eqref{eq:zk_pm} or~\eqref{eq:dW}. It is only necessary to compute $G_k$ if one wants to monitor $\mathcal{L}_\alpha$ during training.

\subsection{Target Loss Functions}
In a small MLP based a sensitivity analysis of the impact of the choice of $\beta_L$ (Section~\ref{sec:mlp_experiments}), we employed a linearized MSE function by using~\eqref{eq:zL_linear}. As mentioned in Section~\ref{sec:inference}, this allowed us to try out larger values of $\beta_L$.
However, in experiments where repeated state updates are made, one needs to make a choice of what linearization point to use at subsequent iterations. For this reason we simply employed the regular MSE loss in the remaining MLP experiments of Section~\ref{sec:mlp_experiments} and used the update rule~\eqref{eq:zL_MSE}.

For the experiments on deep convolutional neural networks in Section~\ref{sec:cnn_experiments} the mean square error loss was insufficient for achieving good performance, so a linearized version of softmax cross-entropy (the cross-entropy loss of the softmax of the output neurons) was employed.

\subsection{Max-Pooling Layers}
As in back-propagation, units participating in a max-pooling operation are suppressed unless they are the local maximum. The suppressed units do not receive feedback from upstream layers and do not propagate their activity forwards. For the efficient version of dual propagation ({\color[HTML]{0077BB}blue} in Algorithm~\ref{alg:dualprop}) this is achieved by using the standard autodiff rules for max-pooling layers. This is similar to the approach taken by~\cite{laborieux2021scaling}.

\section{Experiments}
We evaluate dual propagation on MNIST, CIFAR10, CIFAR100, and Imagenet32x32. MNIST is used to analyse variations of the algorithm, the other datasets for comparison with back-propagation. Through this we show that DP performs nearly identical to BP in both runtime and accuracy. Our code is available on github\footnote{https://github.com/Rasmuskh/Dual-Propagation}.

\subsection{MLP Trained on MNIST}\label{sec:mlp_experiments}
Before applying dual propagation on more challenging tasks and datasets, we explored the impact of variations of the algorithm on the MNIST digit classification dataset, using a ReLU MLP with architecture $784-1000-1000-1000-1000-10$.

Apart from DP and R-DP presented in Algorithm~\ref{alg:dualprop}, we also explore three other variations. Lazy dual propagation (L-DP) differs from DP only in that hidden and output units are not reset to zero activity before processing a new batch of data. This means that feature vectors from previous data points are allowed to provide potentially disruptive feedback. 
In multi-step dual propagation (MS-DP) we perform five inference passes up and down on the same batch with a weight update after each full pass. This is qualitatively similar to the repeated weight updates employed in~\cite{ernoult2020eqprop_continual_weight_updates, salvatori2022incremental}.
Parallel dual propagation (P-DP) updates all neurons in parallel. This method requires $2L-1$ updates for an informative signal to reach all layers. The output neurons only receive feedback from the loss function during the last $L$ updates, since the loss signal isn't meaningful until the input signal has had a chance to reach the output layer.

As mentioned, it is advantageous to choose $\beta_k = \beta_{k+1}$ for $1\leq k \leq L$, but that still leaves us with the freedom to choose $\beta_L$. The error signal arriving at layer $L-1$ is inversely proportional to $\beta_L$, making it necessary to divide the learning rate by $\beta_L$.

The impact of different choices of the hyper-parameter $\beta_L$ is explored in an initial sensitivity experiment. As shown in~Table~\ref{tab:beta_analysis}, $\beta_L=1$ was found to perform the best and was consequently used in subsequent experiments.

\begin{table}[h]
    \centering
    \caption{Performance impact of different choices of $\beta_L$ on MNIST test accuracy.}
    \begin{tabular*}{\linewidth}{llllll}
    \hline
    $\beta_L$     & 0.01  & 0.1   & 1.0   & 10    & 100   \\ \hline
    Test acc. (\%) & 98.04 & 98.34 & 98.38 & 94.95 & 85.42 \\ \hline
    \end{tabular*}\label{tab:beta_analysis}
\end{table}

\begin{table}[h]
    \centering
    \caption{MNIST test accuracies obtained with an MLP with 4 hidden layers (of 1000 units each).}
    \label{tab:mnist}
    \small
    \tabcolsep=0.11cm
    \begin{tabular*}{\linewidth}{llllllll}
    \hline
    Method & BP & DP & MS-DP & L-DP & P-DP & R-DP-100\\ \hline
    Test & 98.45 & 98.43 & 98.40 & 98.42 & 98.47 & 98.48  \\
    acc (\%) & $\pm0.04$ & $\pm0.03$ & $\pm0.02$ & $\pm0.07$ & $\pm0.04$ & $\pm0.11$\\ \hline
    \end{tabular*}
\end{table}

\begin{figure}[h]
    \centering
    \includegraphics[width=0.49\textwidth]{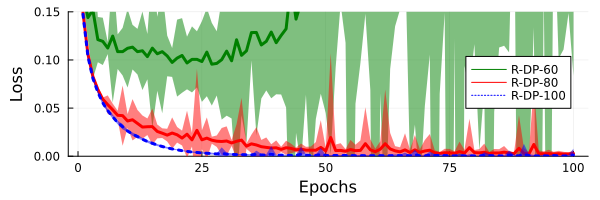}
    \caption{MNIST training loss for dual propagation for different numbers of randomly ordered layer updates. Shaded regions indicate $\pm3$ standard deviations.}
    \label{fig:random_dualprop}
\end{figure}

The MLP was trained on the MNIST dataset using the ADAM optimizer~\cite{kingma2014adam}. 10\% of the training data was reserved for validation, and performance on the validation data was used to select which checkpoint of the model to evaluate on the test dataset. 
The resulting test accuracy, summarized in Table~\ref{tab:mnist}, illustrate that the variations of DP (and BP) essentially perform equivalently. However, for R-DP it was found to be essential that a sufficient number of neuron updates are made. Fig.~\ref{fig:random_dualprop} illustrates this. With 60 updates learning fails, with 80 updates the loss converges noisily, and with 100 updates it converges smoothly.

\begin{figure}[ht]
    \centering
    \includegraphics[width=0.49\textwidth]{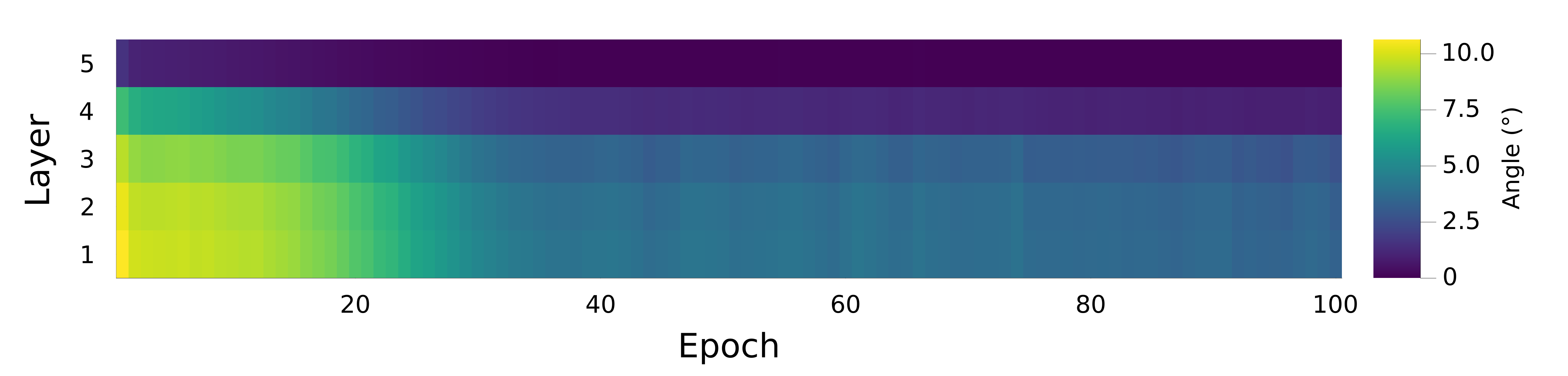}
    \caption{Layerwise angle between gradients computed using dual propagation and back-propagation in a five layer MLP.}
    \label{fig:mnist_grad_cossim}
\end{figure}

Fig.~\ref{fig:mnist_grad_cossim} shows the average angle between the layerwise gradients computed by DP and the corresponding analytical BP gradient for one of the random seeds. Throughout the $100$ epoch run, the average angle remains below $11.5^{\circ}$ (corresponding to a minimum cosine similarity of $0.98$).

\subsection{Deep CNN Experiments}\label{sec:cnn_experiments}
\begin{figure}[tb]
     \centering
     \begin{subfigure}[b]{0.49\textwidth}
         \centering
         \includegraphics[width=\textwidth]{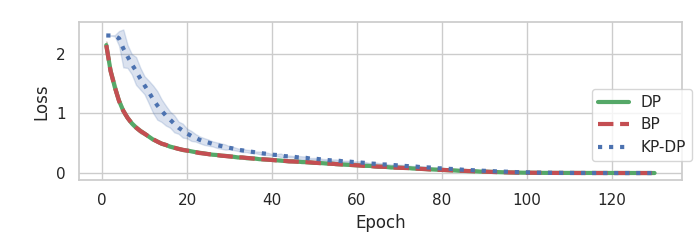}
         \caption{}
         \label{fig:fig:cifar10_train_loss}
     \end{subfigure}
     \hfill
     \begin{subfigure}[b]{0.49\textwidth}
         \centering
         \includegraphics[width=\textwidth]{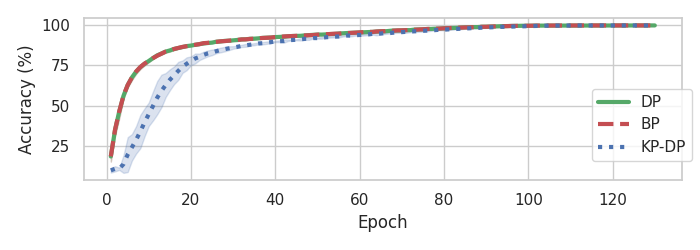}
         \caption{}
         \label{fig:cifar10_train_acc}
     \end{subfigure}
        \caption{Training metrics (accuracy and softmax cross-entropy loss), for VGG16 networks trained on CIFAR10 with back-propagation (BP), dual propagation (DP) and a variant of dual propagation with distinct feedback weights trained with Kolen-Pollack algorithm KP-DP. Shaded regions indicate $\pm3$ standard deviations.}
        \label{fig:cifar10_train}
\end{figure}

We benchmarked dual propagation against back-propagation on the CIFAR10 and CIFAR100 datasets~\cite{krizhevsky2009cifar}. The algorithms were used to train a VGG16 model~\cite{simonyan2014vgg}. A variant model with distinct forward and backward weights was also trained with dual propagation, using the Kolen-Pollack algorithm to learn the feedback weights~\cite{kolenpollack1994}.
The models were trained using standard data augmentation techniques (random-crops and horizontal flips) and the training data of $50\,000$ images was split into $45\,000$ for training and $5\,000$ for validation. For each run, a snapshot of the model was selected based on validation accuracy and later evaluated on test data.
The average test accuracy across five random seeds are listed in Table~\ref{tab:cnn_results}. 
Interestingly the KP-DP approach performs slightly better on CIFAR100, suggesting that the initially unaligned feedback weights might have a regularizing effect.
Training accuracy and softmax cross-entropy loss across epochs are plotted in Fig.~\ref{fig:cifar10_train}. On an NVIDIA A100 GPU both BP and DP had runtimes of ${\sim}3.5$ seconds per epoch and ${\sim}4.5$ seconds per epoch for CIFAR10 and CIFAR100 respectively (a smaller batchsize was used for CIFAR100). KP-DP had an additional overhead of about $1$ second per epoch, presumably due having to update an additional set of weights.

\begin{table*}[]
    \centering
        \caption{CIFAR10, CIFAR100 and ImageNet32x32 test accuracy, obtained with dual propagation (DP) Kolen-Polack dual propagation (KP-DP) and back-propagation (BP) using a VGG16 architecture. (*) The to our knowledge best published results for equilibrium propagation (EP)~\cite{laborieux2022holomorphic} and difference target propagation (DTP)~\cite{ernoult2022} are listed in the rightmost columns for reference (note that these are based on different network architectures).}
        \label{tab:cnn_results}
        \begin{tabular}{lllll||l|l}
        \hline
        Method                    &       &BP                   & DP                & KP-DP             & EP$^*$        & DTP$^*$ \\ \hline
        CIFAR10                   & Top-1 & $92.26\pm0.23$      & $92.30\pm0.11$    & $91.84\pm0.11$    & $88.6\pm0.2$  & $89.38\pm0.20$\\ \hline
        \multirow{2}{*}{CIFAR100} & Top-1 & $69.63\pm0.24$      & $69.57\pm0.51$    & $70.40\pm0.25$    & $61.6\pm0.1$  & $-$    \\
                                  & Top-5 & $88.13\pm0.22$      & $88.36\pm0.13$    & $88.57 \pm 0.15$  & $86.0\pm0.1$  & $-$    \\ \hline
        \multirow{2}{*}{ImageNet 32x32}            & Top-1 & $41.28\pm0.19$      & $41.48\pm0.19$    & $-$               & $36.5\pm0.3$  & $36.81$    \\
                                  & Top-5 & $64.89\pm0.11$      & $64.90\pm0.13$    & $-$               & $60.8\pm0.4$  & $60.54$    \\ \hline
    \end{tabular}
\end{table*}



The VGG16 model was also trained on Imagenet32x32, yielding essentially equivalent performance for DP and BP. Standard data augmentation techniques (random-crops and horizontal flips) was also employed in this experiment. As Imagenet32x32 does not have a public test dataset we used the validation data as test data and reserved 5\% of the training data for validation. Training time per epoch was $\sim$61 seconds on an NVIDIA A100 gpu for both methods.
The average test accuracies accross five random seeds are listed in Table~\ref{tab:cnn_results}.
Kolen-Pollack learning did not work for this dataset. We observe that the asymmetric network became successively more sensitive to the choice of hyper-parameters as the number of classes increased (from CIFAR10 to CIFAR100 to Imagenet32x32).

For reference we also report the, to our knowledge, best results for NGRAD algorithms in the rightmost columns of Table~\ref{tab:cnn_results}, namely the results for (Holomorphic) Equilibrium propagation~(\cite{laborieux2022holomorphic}) and a recent variant of difference target propagation~\cite{ernoult2022}. We emphasize that both EP and DTP are capable of approximating back-propagation, provided a sufficient amount of computational resources and time is dedicated. For EP this means that sufficiently many inference iterations must be run and for DTP it means that sufficiently many iterations of feedback weight learning must be run. Thus, the performance difference between these algorithms and DP, as shown in Table~\ref{tab:cnn_results}, mainly reflects their high computational costs, which makes hyperparameter search challenging and, crucially, makes training very deep networks computationally infeasible. Consequently both the results for EP~\cite{laborieux2022holomorphic} and DTP~\cite{ernoult2022} were obtained using small VGG-like networks (5-7 layers).



\section{Conclusion}

Variations of contrastive Hebbian learning are gaining traction as they are possibly highly suitable for DNN training on energy-efficient analog computing devices. However, their high computational demand when implemented on digital hardware is clearly a disadvantage when exploring these algorithms. 
Our proposed algorithm, dual propagation, differs from traditional contrastive Hebbian learning algorithms in that the errors are computed across different compartments of individual neurons rather than across different temporal states of the same neuron. The resulting formulation allows for closed-form neuron update rules, which makes dual propagation competitive with back-propagation, both in terms of accuracy and runtime.


An important question for future work is whether the proposed dual propagation method is easily implementable on analog computing hardware.
In the digital realm, dual propagation may prove highly valuable for training quantized neural networks by a suitable choice of $\beta_k$, hence steering the finite difference approximation of the possibly non-smooth activation function.


\paragraph{Acknowledgements}
This work was supported by the Wallenberg AI, Autonomous Systems and Software Program (WASP) funded by the Knut and Alice Wallenberg Foundation, and by the Chalmers AI Research Centre (CHAIR).
The experiments were enabled by the supercomputing resource Berzelius provided by National Supercomputer Centre at Linköping University and the Knut and Alice Wallenberg foundation. 

\bibliography{main}
\bibliographystyle{icml2023}

\newpage
\appendix
\onecolumn

\section{Deriving the update relations~\eqref{eq:zk_pm}}

\paragraph{Solving for $z_k^+$:}
The terms in $\cL_\alpha$~\eqref{eq:L_alpha} dependent on $z_k^+$ are given by
\begin{align}
  V_k^+(z_k^+) 
  {} &= \beta_k^{-1} \Paren{ \alpha G_k(z_k^+) - \alpha (z_k^+)\tr W_{k-1} z_{k-1}^+) + \bar\alpha G_k(z_k^+) - \bar\alpha (z_k^+)\tr W_{k-1} z_{k-1}^- } \nonumber \\
  {} &+ \beta_{k+1}^{-1}\alpha \Paren{ G_{k+1}(z_{k+1}^+) - (z_{k+1}^+)\tr W_k z_k^+ - G_{k+1}(z_{k+1}^-) + (z_{k+1}^-)\tr W_k z_k^+ } \nonumber \\
  {} &\doteq \beta_k^{-1} \Paren{ G_k(z_k^+) - \alpha (z_k^+)\tr W_{k-1} z_{k-1}^+) - \bar\alpha (z_k^+)\tr W_{k-1} z_{k-1}^- }
       + \beta_{k+1}^{-1}\alpha \Paren{ - (z_{k+1}^+)\tr W_k z_k^+ + (z_{k+1}^-)\tr W_k z_k^+ } \nonumber \\
  {} &\propto G_k(z_k^+) - \alpha (z_k^+)\tr W_{k-1} z_{k-1}^+) - \bar\alpha (z_k^+)\tr W_{k-1} z_{k-1}^-
       + \tfrac{\alpha\beta_k}{\beta_{k+1}} \Paren{ - (z_{k+1}^+)\tr W_k z_k^+ + (z_{k+1}^-)\tr W_k z_k^+ } \nonumber \\
  {} &= G_k(z_k^+) - (z_k^+)\tr \Paren{ \alpha W_{k-1} z_{k-1}^+  + \bar\alpha W_{k-1} z_{k-1}^- + \tfrac{\alpha\beta_k}{\beta_{k+1}} W_k\tr (z_{k+1}^+ - z_{k+1}^-) }.
\end{align}
Hence, $z_k^+$ is given by
\begin{align}
  z_k^+ \gets f_k\Paren{ \alpha W_{k-1} z_{k-1}^+  + \bar\alpha W_{k-1} z_{k-1}^- + \tfrac{\alpha\beta_k}{\beta_{k+1}} W_k\tr (z_{k+1}^+ - z_{k+1}^-) }.
\end{align}

\paragraph{Solving for $z_k^-$:}
Analogously, the terms in $\cL_\alpha$ dependent on $z_k^-$ are given by
\begin{align}
  V_k^-(z_k^-) 
  {} &\propto -G_k(z_k^-) + \alpha (z_k^-)\tr W_{k-1} z_{k-1}^+ + \bar\alpha (z_k^-)\tr W_{k-1} z_{k-1}^- \nonumber \\
  {} &+ \tfrac{\bar\alpha\beta_k}{\beta_{k+1}} \Paren{ G_{k+1}(z_{k+1}^+) - (z_{k+1}^+)\tr W_k z_k^- - G_{k+1}(z_{k+1}^-) + (z_{k+1}^-)\tr W_k z_k^- } \nonumber \\
  {} &\doteq -G_k(z_k^-) + \alpha (z_k^-)\tr W_{k-1} z_{k-1}^+ + \bar\alpha (z_k^-)\tr W_{k-1} z_{k-1}^- 
       + \tfrac{\bar\alpha\beta_k}{\beta_{k+1}} \Paren{ - (z_{k+1}^+)\tr W_k z_k^- + (z_{k+1}^-)\tr W_k z_k^- } \nonumber \\
  {} &= -G_k(z_k^-) + (z_k^-)\tr \Paren{ \alpha W_{k-1} z_{k-1}^+ + \bar\alpha W_{k-1} z_{k-1}^- + \tfrac{\bar\alpha\beta_k}{\beta_{k+1}} W_k\tr (z_{k+1}^- - z_{k+1}^+) },
\end{align}
which implies
\begin{align}
  z_k^- \gets f_k\Paren{ \alpha W_{k-1} z_{k-1}^+ + \bar\alpha W_{k-1} z_{k-1}^- + \tfrac{\bar\alpha\beta_k}{\beta_{k+1}} W_k\tr (z_{k+1}^- - z_{k+1}^+) }.
\end{align}

\section{Propagation of asymmetric finite differences}
\label{sec:asymmetric}

We consider a more general version of~\eqref{eq:delta_k}:
$\delta_L^\pm$ are given by $\delta_L^+ \gets -\alpha\beta g$ and
$\delta_L^- \gets \bar\alpha \beta g$ (assuming a linearized target loss
$\ell$ with gradient $g$ at the current linearization point), and the two
feedback signals are propagated through the network via
\begin{align}
  \delta_k^+ &\gets f_k\Paren{ W_{k-1}z_{k-1}^* + \alpha W_k\tr \delta_{k+1}^+ } - z_k^*
  & \delta_k^- &\gets z_k^* - f_k\Paren{ W_{k-1}z_{k-1}^* - \bar\alpha W_k\tr \delta_{k+1}^- },
\label{eq:delta_k1}
\end{align}
where $\alpha\in [0,1]$ and $\bar\alpha := 1-\alpha$. Our aim to is
reparametrize $\delta_k^\pm$ using $z_k^\pm$ such that
\begin{align}
  z_k^+-z_k^- = \delta_k^+ + \delta_k^-
  = f_k\Paren{ W_{k-1}z_{k-1}^* + \alpha W_k\tr \delta_{k+1}^+ } - f_k\Paren{ W_{k-1}z_{k-1}^* - \bar\alpha W_k\tr \delta_{k+1}^- }
\end{align}
and $z_k^* = \bar z_k = \alpha z_k^+ + \bar\alpha z_k^-$. After introducing
$\delta_k := \delta_k^+ + \delta_k^-$, combining these constraints yields
\begin{align}
  \alpha z_k^+ = \alpha z_k^- + \alpha \delta_k & & \alpha z_k^+ = z_k^* - \bar\alpha z_k^-,
\end{align}
which implies
\begin{align}
  \alpha z_k^- + \alpha\delta_k - z_k^* + \bar\alpha z_k^- = 0 \iff z_k^- = z_k^* - \alpha\delta_k
\end{align}
and analogously $z_k^+ = z_k^*+\bar\alpha\delta_k$.
Observe that there is an asymmetry in the
role of $\alpha$ in the forward and backward (adjoint) process, e.g.\ choosing
$\alpha=0$ yields $(z_k^+,z_k^-)=(z_k^*-\delta_k,z_k^*)$ and
$(\delta_k^+,\delta_k^-)=(0,\delta_k)$. We have agreement of this model with
the updates in~\eqref{eq:zk_pm} and~\eqref{eq:zL_linear} only when
$\alpha=1/2$. Consequently, the reasoning presented in Section~\ref{sec:analysis}
applies only for the choice of $\alpha=1/2$.

\section{Hyper-parameter settings}
The hyper-parameters used in the experiments of Section~\ref{sec:mlp_experiments} are listed in Table~\ref{tab:mnist_hyper-parameters}. All experiments used these hyper-parameters except for MS-DP which used a learning rate of 6e-6 (to account for the higher number of weight updates per minibatch) and the experiments listed in Table~\ref{tab:beta_analysis}, which only trained for 50 epochs. $\eta$ is the learning rate and $b_1$, $b_2$ and $\epsilon$ are parameters for the ADAM optimizer. The hyper-parameters used in the experiments of Section~\ref{sec:cnn_experiments} are listed in Table~\ref{tab:cifar_hyper-parameters}. For these experiments a linear learning rate warmup schedule was employed followed by cosine decay.
\begin{table}[h]
\centering
\caption{Hyper-parameters used in the MLP experiments.}\label{tab:mnist_hyper-parameters}
\begin{tabular}{|l|lllllll|}
\hline
Dataset & Optimizer & $\eta$ & $b_1$ & $b_2$   & $\epsilon$ & Epochs & batchsize \\ \hline
MNIST & ADAM      & 3e-5 & $0.9$ & $0.999$ & $1e-8$     & 100    & 100       \\ \hline
\end{tabular}
\end{table}

\begin{table}[h]
\footnotesize
\tabcolsep=0.11cm
\centering
\caption{Hyper-parameters used in VGG16 experiments.}
\label{tab:cifar_hyper-parameters}
\begin{tabular}{|l|lllllllll|}
\hline
Dataset & Model & Momentum & $\eta_{start}$ & $\eta_{peak}$ & $\eta_{end}$ & Warmup epochs & Epochs & Weight decay & batchsize \\ \hline
\multirow{3}{*}{CIFAR10} & BP & 0.9 & 0.005 & 0.025 & 0 & 10 & 130 & 5e-4 & 100 \\ \cline{2-10} 
 & DP & 0.9 & 0.005 & 0.025 & 0 & 10 & 130 & 5e-4 & 100 \\ \cline{2-10} 
 & KP-DP & 0.9 & 0.0001 & 0.025 & 0 & 15 & 130 & 5e-4 & 100 \\ \hline
\multirow{3}{*}{CIFAR100} & BP & 0.9 & 0.005 & 0.015 & 0 & 10 & 200 & 5e-4 & 50 \\ \cline{2-10} 
 & DP & 0.9 & 0.005 & 0.015 & 0 & 10 & 200 & 5e-4 & 50 \\ \cline{2-10} 
 & KP-DP & 0.9 & 0.0001 & 0.015 & 0 & 30 & 200 & 5e-4 & 50 \\ \hline
 \multirow{2}{*}{Imagenet32x32} & BP & 0.9 & 0.005 & 0.015 & 0 & 10 & 200 & 5e-4 & 250 \\ \cline{2-10} 
 & DP & 0.9 & 0.005 & 0.015 & 0 & 10 & 200 & 5e-4 & 250 \\ \hline
\end{tabular}
\end{table}

\section{Plots of training metrics}

\begin{figure}[htb]
     \centering
     \begin{subfigure}[b]{0.9\textwidth}
         \begin{subfigure}[b]{0.48\textwidth}
             \centering
             \includegraphics[width=\textwidth]{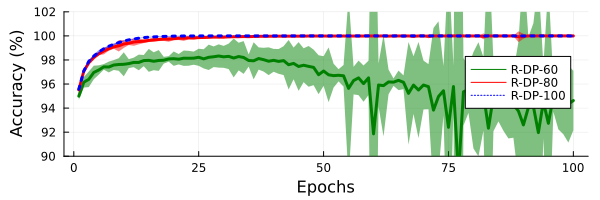}
             \caption{MNIST: Training accuracy.}
             \label{fig:app:mnist_train_acc}
         \end{subfigure}
         \hfill
         \begin{subfigure}[b]{0.48\textwidth}
             \centering
             \includegraphics[width=\textwidth]{images/mnist/R_DP_mean_train_loss.png}
             \caption{MNIST: Training loss.}
             \label{fig:app:mnist_train_loss}
         \end{subfigure}
         \hfill
         \begin{subfigure}[b]{0.48\textwidth}
             \centering
             \includegraphics[width=\textwidth]{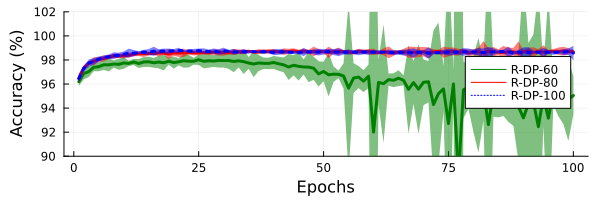}
             \caption{MNIST: Validation accuracy.}
             \label{fig:app:mnist_val_acc}
         \end{subfigure}
         \hfill
         \begin{subfigure}[b]{0.48\textwidth}
             \centering
             \includegraphics[width=\textwidth]{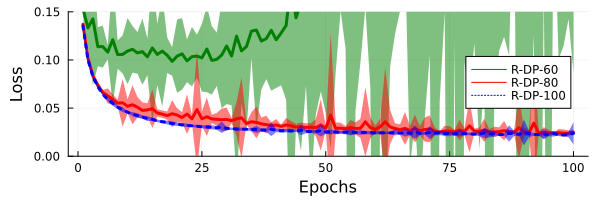}
             \caption{MNIST: Validation loss.}
             \label{fig:app:mnist_val_loss}
         \end{subfigure}
     \end{subfigure}
    \caption{Training metrics (accuracy and MSE), for five layer MLP trained with R-DP using different numbers of random updates. Shaded regions indicate $\pm3$ standard deviations.}
        \label{fig:app:mnist_metrics}
\end{figure}

\begin{figure}[htb]
     \centering
     \begin{subfigure}[b]{0.9\textwidth}
         \begin{subfigure}[b]{0.48\textwidth}
             \centering
             \includegraphics[width=\textwidth]{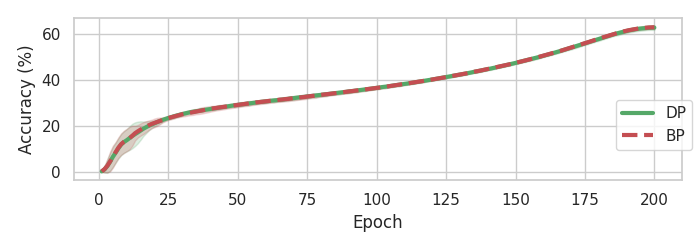}
             \caption{Imagenet32x32: Training accuracy.}
             \label{fig:app:imagenet32x32_train_acc}
         \end{subfigure}
         \hfill
         \begin{subfigure}[b]{0.48\textwidth}
             \centering
             \includegraphics[width=\textwidth]{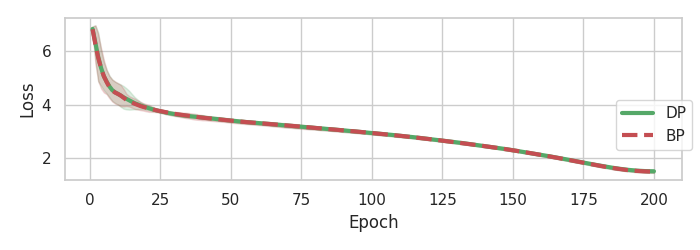}
             \caption{Imagenet32x32: Training loss.}
             \label{fig:app:imagenet32x32_train_loss}
         \end{subfigure}
         \hfill
         \begin{subfigure}[b]{0.48\textwidth}
             \centering
             \includegraphics[width=\textwidth]{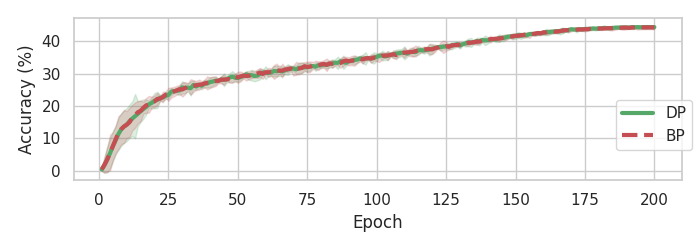}
             \caption{Imagenet32x32: Validation accuracy.}
             \label{fig:app:imagenet32x32_val_acc}
         \end{subfigure}
         \hfill
         \begin{subfigure}[b]{0.48\textwidth}
             \centering
             \includegraphics[width=\textwidth]{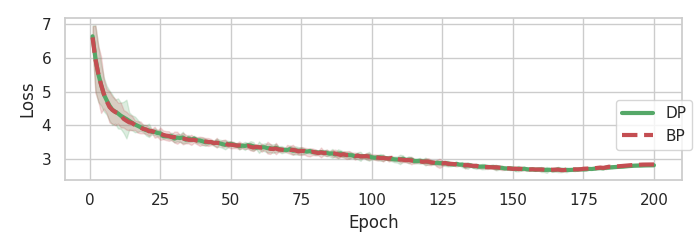}
             \caption{Imagenet32x32: Validation loss.}
             \label{fig:app:imagenet32x32_val_loss}
         \end{subfigure}
    \end{subfigure}
    \caption{Training metrics (accuracy and softmax cross-entropy loss), for VGG16 networks trained with back-propagation (BP) and dual propagation (DP). Shaded regions indicate $\pm3$ standard deviations.}
    \label{fig:imagenet32x32_metrics}
\end{figure}

\begin{figure}[p]
     \centering
     \begin{subfigure}[b]{0.48\textwidth}
         \centering
         \includegraphics[width=\textwidth]{images/cifar10/cifar10_vgg16_acc_train.png}
         \caption{CIFAR10: Training accuracy.}
         \label{fig:app:cifar10_train_acc}
     \end{subfigure}
     \hfill
     \begin{subfigure}[b]{0.48\textwidth}
         \centering
         \includegraphics[width=\textwidth]{images/cifar10/cifar10_vgg16_loss_train.png}
         \caption{CIFAR10: Training loss.}
         \label{fig:app:cifar10_train_loss}
     \end{subfigure}
     \hfill
     \begin{subfigure}[b]{0.48\textwidth}
         \centering
         \includegraphics[width=\textwidth]{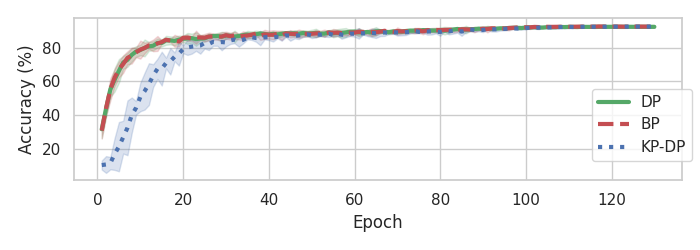}
         \caption{CIFAR10: Validation accuracy.}
         \label{fig:app:cifar10_val_acc}
     \end{subfigure}
     \hfill
     \begin{subfigure}[b]{0.48\textwidth}
         \centering
         \includegraphics[width=\textwidth]{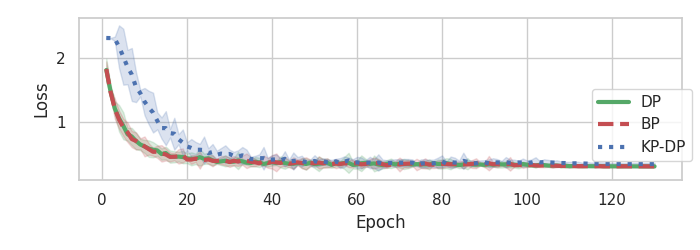}
         \caption{CIFAR10: Validation loss.}
         \label{fig:app:cifar10_val_loss}
     \end{subfigure}
     \hfill
     \begin{subfigure}[b]{0.48\textwidth}
         \centering
         \includegraphics[width=\textwidth]{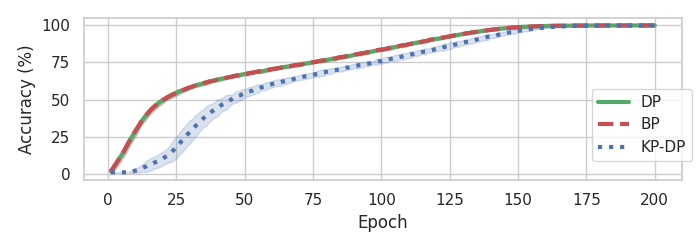}
         \caption{CIFAR100: Training accuracy.}
         \label{fig:app:cifar100_train_acc}
     \end{subfigure}
     \hfill
     \begin{subfigure}[b]{0.48\textwidth}
         \centering
         \includegraphics[width=\textwidth]{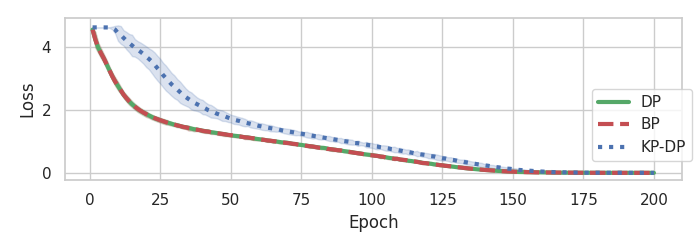}
         \caption{CIFAR100: Training loss.}
         \label{fig:app:cifar100_train_loss}
     \end{subfigure}
     \hfill
          \begin{subfigure}[b]{0.48\textwidth}
         \centering
         \includegraphics[width=\textwidth]{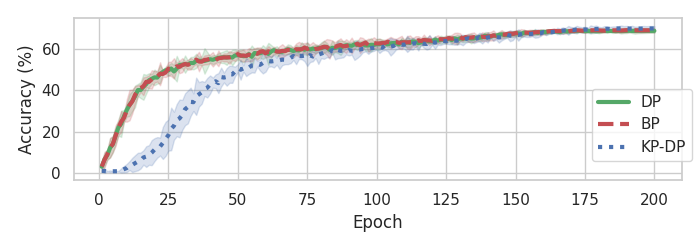}
         \caption{CIFAR100: Validation accuracy.}
         \label{fig:app:cifar100_val_acc}
     \end{subfigure}
     \hfill
     \begin{subfigure}[b]{0.48\textwidth}
         \centering
         \includegraphics[width=\textwidth]{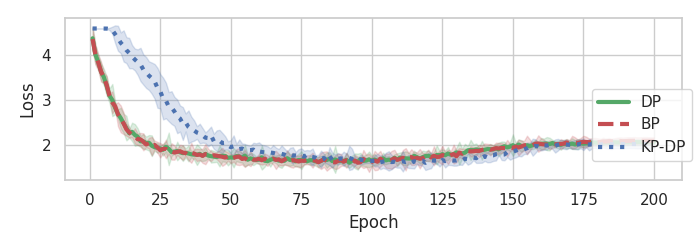}
         \caption{CIFAR100: Validation loss.}
         \label{fig:app:cifar100_val_loss}
     \end{subfigure}

    \caption{Training metrics (accuracy and softmax cross-entropy loss), for VGG16 networks trained with back-propagation (BP), dual propagation (DP) and a variant of dual propagation with distinct feedback weights trained with Kolen-Pollack algorithm KP-DP. Shaded regions indicate $\pm3$ standard deviations.}
        \label{fig:app:cifar_metrics}
\end{figure}

We illustrate the training evolution of DP and BP on the various datasets in Figs.~\ref{fig:app:mnist_metrics},~\ref{fig:imagenet32x32_metrics} and~\ref{fig:app:cifar_metrics}, respectively.

\section{Network architecture}
To allow for full reproducibility, Table~\ref{tab:vgg16} shows the architecture of the VGG16 version without batchnorm used in Sec~\ref{sec:cnn_experiments}.

\begin{table}[p]
\centering
\caption{The architecture of the VGG16 network used in CIFAR10, CIFAR100 and Imagenet32x32 experiments. All convolutional layers have $3\times3$ kernels with stride $1$, all pooling layers are of size $2\times2$ with stride $2$. I.e., convolutional layers preserve image size, pooling layers half it. The $z^+/z^-$ dyad is implemented in the ReLU layers. For the Kolen-Pollack variation, all Convolution and Fully Connected layers are replaced with asymmetric variants.}
\footnotesize
\label{tab:vgg16}
\begin{tabular}{r|ccc}
           & Type            & Kernel/Stride & Channels or Size \\
\hline\hline
\rownumber & Convolutional   & $3\times3$ / $1$ & $64$  \\
\rownumber & ReLU            & --               & --    \\
\rownumber & Convolutional   & $3\times3$ / $1$ & $64$  \\
\rownumber & ReLU            & --               & --    \\
\rownumber & Max Pool        & $2\times2$ / $2$ &       \\
\rownumber & Convolutional   & $3\times3$ / $1$ & $128$ \\
\rownumber & ReLU            & --               & --    \\
\rownumber & Convolutional   & $3\times3$ / $1$ & $128$ \\
\rownumber & ReLU            & --               & --    \\
\rownumber & Max Pool        & $2\times2$ / $2$ &       \\
\rownumber & Convolutional   & $3\times3$ / $1$ & $256$ \\
\rownumber & ReLU            & --               & --    \\
\rownumber & Convolutional   & $3\times3$ / $1$ & $256$ \\
\rownumber & ReLU            & --               & --    \\
\rownumber & Convolutional   & $3\times3$ / $1$ & $256$ \\
\rownumber & ReLU            & --               & --    \\
\rownumber & Max Pool        & $2\times2$ / $2$ & --    \\
\rownumber & Convolutional   & $3\times3$ / $1$ & $512$ \\
\rownumber & ReLU            & --               & --    \\
\rownumber & Convolutional   & $3\times3$ / $1$ & $512$ \\
\rownumber & ReLU            & --               & --    \\
\rownumber & Convolutional   & $3\times3$ / $1$ & $512$ \\
\rownumber & ReLU            & --               & --    \\
\rownumber & Max Pool        & $2\times2$ / $2$ & --    \\
\rownumber & Convolutional   & $3\times3$ / $1$ & $512$ \\
\rownumber & ReLU            & --               & --    \\
\rownumber & Convolutional   & $3\times3$ / $1$ & $512$ \\
\rownumber & ReLU            & --               & --    \\
\rownumber & Convolutional   & $3\times3$ / $1$ & $512$ \\
\rownumber & ReLU            & --               & --    \\
\rownumber & Max Pool        & $2\times2$ / $2$ & --    \\
\rownumber & Flatten         & --               & --    \\
\rownumber & Fully Connected & --               & $4096$\\
\rownumber & ReLU            & --               & --    \\
\rownumber & Fully Connected & --               & $4096$\\
\rownumber & ReLU            & --               & --    \\
\rownumber & Fully Connected & --               & \#Classes\\
\end{tabular}
\end{table}

\end{document}